\documentclass[11pt,a4paper]{article}
\usepackage[hyperref]{emnlp-ijcnlp-2019}
\usepackage{times}
\usepackage{latexsym}

\usepackage{url}

\aclfinalcopy 

\usepackage{multirow}
\usepackage{bm}

\usepackage{enumitem}
\usepackage{amsfonts}
\usepackage{amssymb}

\usepackage{amsmath}
\usepackage{graphicx}
\usepackage{color}
\usepackage{subcaption}
\definecolor{newblue}{RGB}{70,118,214}
\definecolor{neworange}{RGB}{206,151,90}
\definecolor{newpurple}{RGB}{131,106,212}
\definecolor{newgreen}{RGB}{58,105,57}

\definecolor{Orange}{rgb}{1,0.5,0}

\def\argmax{\mathop{\rm argmax}}

\title{Sequence-to-sequence Pre-training with Data Augmentation \\for Sentence Rewriting}

\author{
Yi Zhang$^{1}$\thanks{\ \ This work was done during the first author's internship at Microsoft Research Asia.}~\thanks{~~Equal contribution} ~~~ Tao Ge$^{2}$\footnotemark[2] ~~~~ Furu Wei$^2$ ~~~ Ming Zhou$^2$ ~~~ Xu Sun$^1$\\
$^1$Peking University, Beijing, China\\
$^2$Microsoft Research Asia, Beijing, China\\
{\tt \{zhangyi16, xusun\}@pku.edu.cn} \\
{\tt \{tage, fuwei, mingzhou\}@microsoft.com}}

\date{}

\usepackage{dcolumn}
\newcolumntype{d}[1]{D{.}{.}{#1}}

\begin{document}
\maketitle
\begin{abstract}

We study sequence-to-sequence (seq2seq) pre-training with data augmentation for sentence rewriting. Instead of training a seq2seq model with gold training data and augmented data simultaneously, we separate them to train in different phases: pre-training with the augmented data and fine-tuning with the gold data. We also introduce multiple data augmentation methods to help model pre-training for sentence rewriting.
We evaluate our approach in two typical well-defined sentence rewriting tasks: \textbf{G}rammatical \textbf{E}rror \textbf{C}orrection (GEC) and \textbf{F}ormality \textbf{S}tyle \textbf{T}ransfer (FST). Experiments demonstrate our approach can better utilize augmented data without hurting the model's trust in gold data and further improve the model's performance with our proposed data augmentation methods. 

Our approach substantially advances the state-of-the-art results in well-recognized sentence rewriting benchmarks over both GEC and FST.  Specifically, it pushes the CoNLL-2014 benchmark's $F_{0.5}$ score and JFLEG Test GLEU score to \textbf{62.61} and \textbf{63.54} in the restricted training setting, \textbf{66.77} and \textbf{65.22} respectively in the unrestricted setting, and advances GYAFC benchmark's BLEU to \textbf{74.24} (\textbf{2.23} absolute improvement) in E\&M domain and \textbf{77.97} (\textbf{2.64} absolute improvement) in F\&R domain.


\end{abstract}

\section{Introduction}

\begin{figure}[t]
	\centering
    \includegraphics[width=7.5cm]{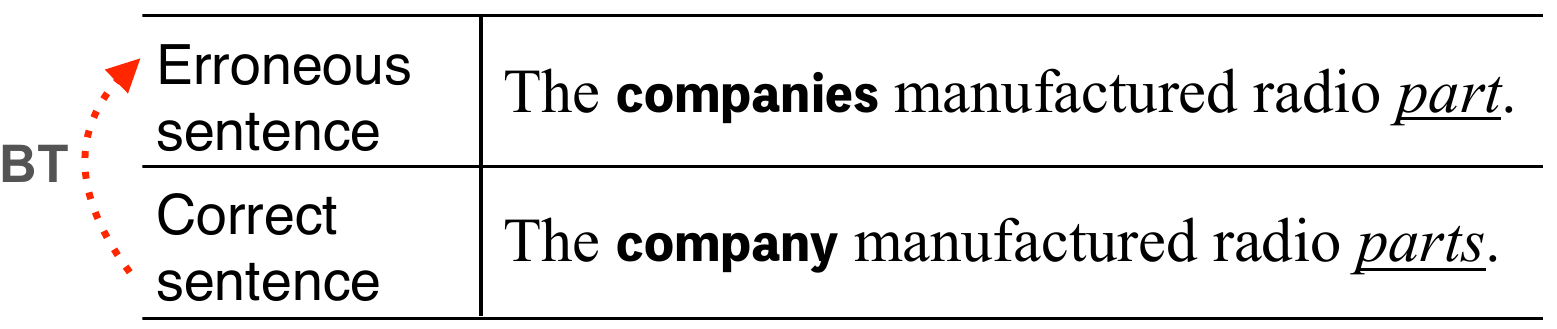}\vspace{-0.15cm}
    \caption{An augmented sentence pair generated through back-translation (BT) for GEC. Though it includes useful rewriting knowledge (the underlined text) for GEC, it additionally introduces undesirable edit (the bold text) which may lead the model to learn to rewrite the content that should not be edited. }\vspace{-0.35cm}
    \label{bad_case}
\label{example}
	\vspace{-0.2cm}
\end{figure}

Data augmentation proves effective in alleviating the issue of insufficient training data because it can help improve the model's generalization ability and reduce the risk of overfitting. For sequence-to-sequence (seq2seq) learning in Natural Language Processing (NLP) tasks, previous studies ~\cite{DBLP:conf/acl/SennrichHB16,edunov2018understanding,DBLP:journals/mt/KarakantaDG18,DBLP:journals/corr/abs-1808-07512} using data augmentation tend to train with the gold data and augmented data simultaneously. Despite the certain effectiveness of the approaches, we find that they suffer from a limitation when applied to sentence rewriting: since simultaneous training does not discriminate between gold and augmented data, the noisy, unnecessary and even erroneous edits introduced in the augmented data tend to make the model become aggressive to rewrite the content that should not be edited, as Figure \ref{bad_case} shows, which is undesirable for sentence rewriting.

To address the issue for better utilizing the augmented data for seq2seq learning in sentence rewriting, we study seq2seq pre-training with data augmentation. 
Instead of training with gold and augmented data simultaneously, our approach trains the model with augmented and gold data in two phases: pre-training and fine-tuning. In the pre-training phase, we train a seq2seq model from scratch with augmented data to help the model learn contextualized representation (encoding), sentence generation (decoding) and potentially useful transformation knowledge (mapping from the source to the target); while in the fine-tuning phase, the model can fully concentrate on the gold training data. In contrast to the previous approaches that train the model with gold and augmented in the same phase, our approach can not only learn useful information from the augmented data, but also avoid the risk of being overwhelmed and adversely affected by the augmented data.

Moreover, we introduce three data augmentation methods to help seq2seq pre-training for sentence rewriting: back translation, feature discrimination and multi-task transfer, which are helpful in improving the model's generalization ability, and also introduce additional rewriting knowledge, as depicted in Figure \ref{example}.

We evaluate our approach in two typical well-defined sentence rewriting tasks: \textbf{G}rammatical \textbf{E}rror \textbf{C}orrection (GEC) and \textbf{F}ormality \textbf{S}tyle \textbf{T}ransfer (FST). Experiments show our approach is more effective to utilize the various augmented data and significantly improves the model, obtaining the state-of-the-art results in both of the tasks.

\begin{figure}[t]
	\centering
    \includegraphics[width=7.5cm]{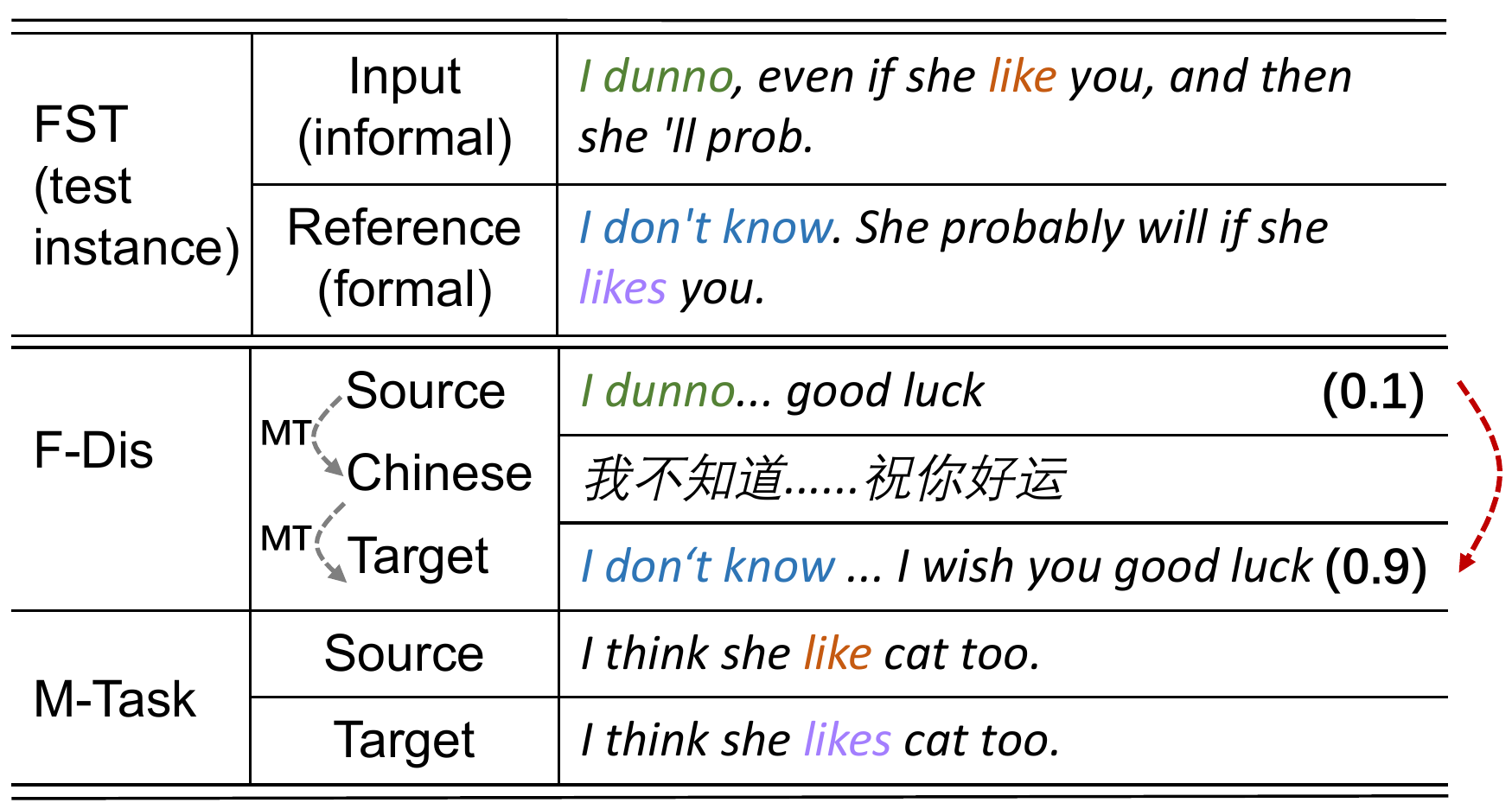}\vspace{-0.2cm}
	\caption{An example that \textbf{F}ormality \textbf{S}tyle \textbf{T}ransfer (FST) benefits from data augmented via \textbf{f}eature \textbf{dis}crimination (\textbf{F-Dis}) and \textbf{m}ulti-\textbf{task} transfer (\textbf{M-Task}). F-Dis identifies useful sentence pairs whose target's formality score (the numbers in the parentheses) is higher than the source, from paraphrase sentences generated by cross-lingual MT, while M-Task utilizes training data for GEC to help formality improvement.}
	\label{example}
	\vspace{-0.2cm}
\end{figure}

Our contributions are summarized as follows:

\begin{itemize}
    \item We study seq2seq pre-training with data augmentation by empirically comparing with other training paradigms for sentence rewriting, confirming its effectiveness and advantages for sentence rewriting tasks.  
    \item We introduce multiple data augmentation ideas for sentence rewriting, which can improve the quality and diversity of the augmented data and introduce additional rewriting knowledge to benefit model pre-training.
    \item Our approach substantially advances the state-of-the-art in all the three important benchmarks (CoNLL2014, JFLEG, GYAFC) in GEC and FST sentence rewriting tasks.
\end{itemize}

\section{Background}

\subsection{Sequence-to-sequence learning}
Sequence-to-sequence (seq2seq) \cite{DBLP:conf/nips/SutskeverVL14,DBLP:journals/corr/ChoMGBSB14} learning has achieved tremendous success in various NLP tasks. Given a source sentence $\boldsymbol{x}$, a seq2seq model learns to generate its target sentence $\boldsymbol{y}$. 
The model is usually trained by maximizing the log-likelihood of the training source-target sentence pairs:
 \begin{equation}
 \small
 \hat{\bm{\theta}} = \argmax_{\bm{\theta}}\sum_{(\bm{x}, \bm{y}) \in \mathcal{T}} \log p(\bm y|\bm x; \bm{\theta})
 \end{equation}
where $\mathcal{T}$ denotes the training set (i.e., source-target parallel sentence pairs) and $\bm{\theta}$ denotes the parameters of the model.

During inference, the decoder generates output $\boldsymbol{y}$ autoregressively by maximizing $p(\bm y|\bm x; \hat{\bm{\theta}})$:
\vspace{-0.2cm}
 \begin{equation}
 \small
 p(\bm y|\bm x; \hat{\bm{\theta}}) = \prod_{i=1}^L p(y_i| \bm x, \bm{y}_{<i}; \hat{\bm{\theta}}) 
 \end{equation}

\subsection{Data Augmentation}
To train a good-performing neural network, sufficient training data is indispensable. However, most tasks lack the annotated training data. As a result, the model may suffer from unsatisfactory generalization ability, as well as robustness defending perturbations outside the training data.

To improve the model's generalization ability, data augmentation is employed to enrich the training set with additional augmented data which is usually artificially generated:
\vspace{-0.1cm}
\begin{equation}
    \mathcal{T} = \mathcal{T}_{orig} \cup \mathcal{T}_{aug}
\end{equation}
\noindent where $\mathcal{T}_{orig}$ and $\mathcal{T}_{aug}$ denote the original training set and the augmented training data respectively. 

\section{Approach}
\subsection{Seq2seq Pre-training \& Fine-tuning}

\begin{figure}[t]
	\centering
	\subcaptionbox{Simultaneous Training \label{pre-train}}
	{\includegraphics[width=0.45\linewidth]{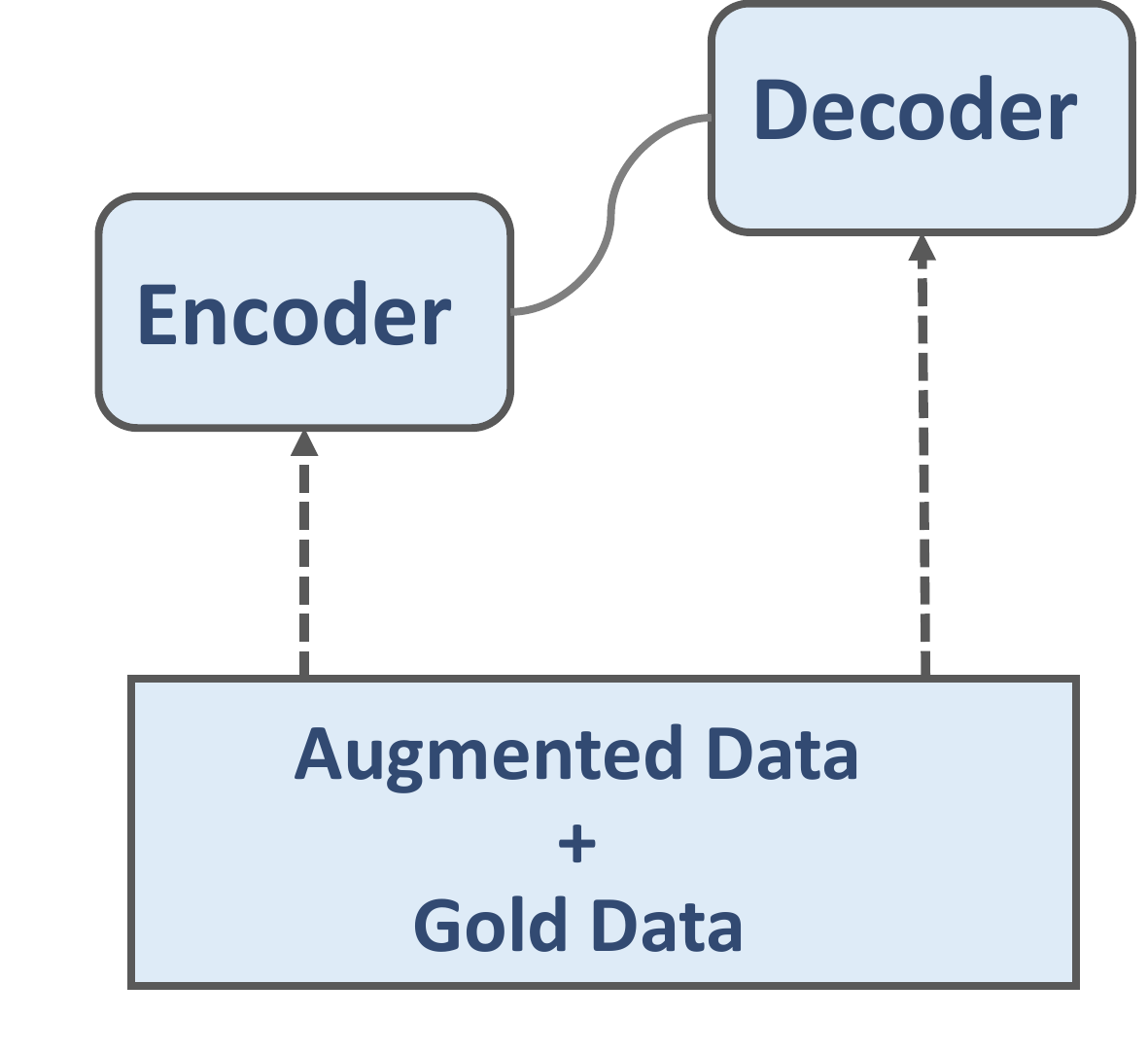}} 
	\subcaptionbox{Pre-training \& Fine-tuning \label{tune}}
	{\includegraphics[width=0.5\linewidth]{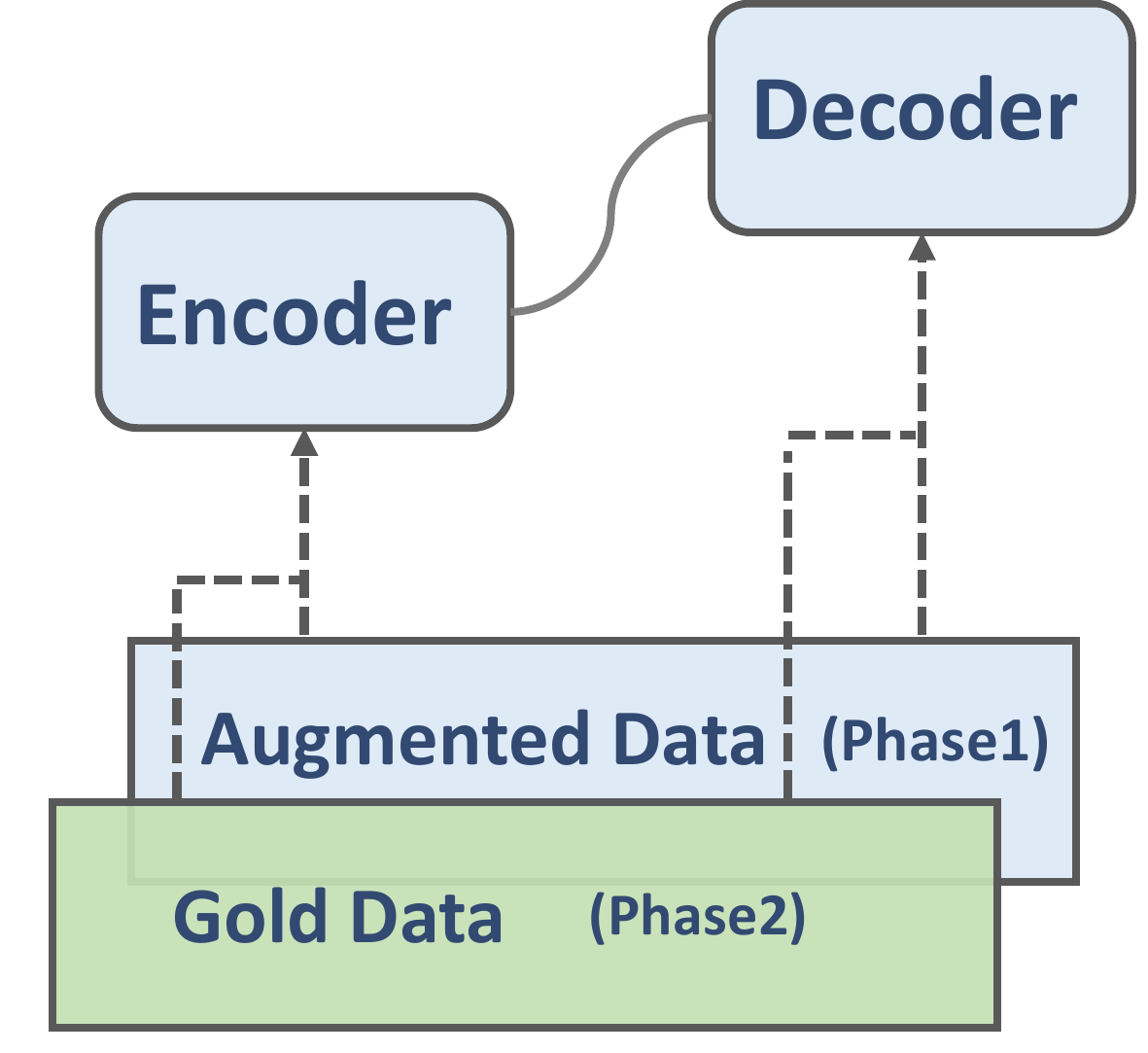}}\vspace{-0.4cm}
	\caption{Comparison between (a) Simultaneous Training and (b) Pre-training \& Fine-tuning framework.}
	\label{fig:pt-ft}
	\vspace{-0.2cm}
\end{figure}

In general, massive augmented data can help a seq2seq model to learn contextualized representations, sentence generation and source-target alignments. However, it is usually noisier and less valuable than gold training data. In simultaneous training (Figure \ref{fig:pt-ft}(a)), the massive augmented data tends to overwhelm the gold data and introduce unnecessary and even erroneous editing knowledge, which is undesirable for sentence rewriting.

To better exploit the augmented data, we propose to first pre-train the model with augmented data and then fine-tune the model with gold training data (Figure \ref{fig:pt-ft}(b)). In our pre-training \& fine-tuning approach, the augmented data is not treated equally to the gold data; instead it only serves as prior knowledge that can be updated and even overwritten during the fine-tuning phase. Then the model can better learn from the gold data without being overwhelmed or distracted by the augmented data. Moreover, separating the augmented and gold data into different training phases makes the model become more tolerant to noise in augmented data, which reduces the quality requirement for the augmented data and enables the model to use noisier augmented data and even training data from other tasks (See Section \ref{subsubsec:multi-task}).

\subsection{Data Augmentation for Text Rewriting}\label{subsec:augmentation}
We study three data augmentation methods for seq2seq sentence rewriting: back translation (Section \ref{subsubsec:bt}), feature discrimination (Section \ref{subsubsec:f-dis}) and multi-task transfer (Section \ref{subsubsec:multi-task}).

\subsubsection{Back translation}\label{subsubsec:bt}
The original idea of back translation \cite{DBLP:conf/acl/SennrichHB16} is to train a target-to-source seq2seq model using bi-lingual parallel corpora, and use the model to generate source language sentences from target monolingual sentences, establishing synthetic parallel sentence pairs.

Although back translation is originally proposed for machine translation (MT), it can be easily generalized to sentence rewriting tasks where source and target are in the same language. In this paper, we use back translation as our basic data augmentation method.

\subsubsection{Feature discrimination}\label{subsubsec:f-dis}

For a well-defined sentence rewriting task, the target sentence is usually expected to improve the source sentence in some aspects without changing its meaning. For instance, for GEC, the target sentence should be grammatically correct, more fluent and native-sounding than the source sentence; while for FST, the target sentence should look more formal than the source sentence. With this motivation, we propose feature discrimination which identifies valuable sentence pairs using a feature-based discriminator from paraphrased sentences for a specific sentence rewriting task.

To make it easy to understand, we present two examples of feature discrimination for data augmentation in GEC and FST respectively.

\begin{figure}[t]
    \centering
    \includegraphics[width=7.5cm]{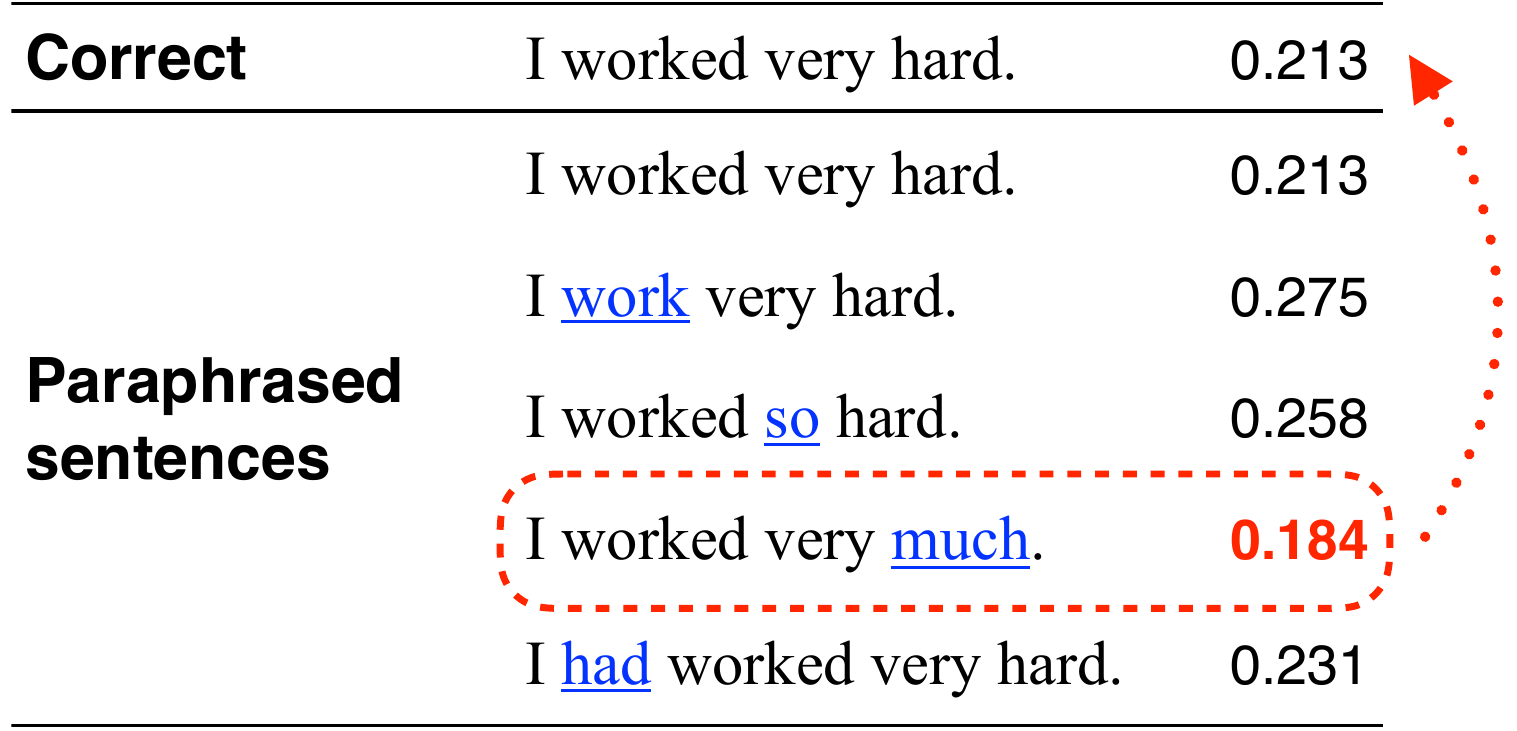}
    \caption{Fluency discrimination for GEC. The paraphrased sentences are generated by a back translation model trained on GEC parallel data. The scores right after the sentences are their fluency scores. The fluency discriminator only chooses the sentence whose fluency score is lower than the correct sentence and pairs them (by the red dashed arrow) as augmented data.}\vspace{-0.35cm}
    \label{fig:fb}
\end{figure}

\subsubsection*{GEC: Fluency discrimination}
For GEC, data augmentation establishes new parallel sentence pairs through deriving a source sentence with grammatical errors by making small modifications to a correct sentence. However, the derived source sentence sometimes does not have grammatical issues; instead, it is just a paraphrased sentence to the correct sentence. Training with such pairs will make the model prone to edit a sentence even if the sentence has no grammatical issues, which may introduce unnecessary and undesirable edits, as depicted in Figure \ref{fig:fb}.

To address this challenge, we use \textit{fluency discrimination}, whose idea was first proposed by \newcite{ge2018fluency}. Fluency discrimination is motivated by that the target correct sentence should be more fluent than the source sentence. It uses a discriminator to evaluate sentences' fluency defined in Eq (\ref{eq:fluency}) by \citet{ge2018fluency}, and only chooses the sentence whose fluency score is lower than the correct sentence, and then pairs them as augmented data. In this way, the undesirable sentence pairs for GEC can be filtered out, as shown in Figure \ref{fig:fb}.
\vspace{-1\baselineskip}

\begin{small}
\begin{align}\label{eq:fluency}
& f(\boldsymbol{x})=\frac{1}{1+H(\boldsymbol{x})} \\
 H(\boldsymbol{x})&=-\frac{\sum_{i=1}^{|\boldsymbol{x}|}\log P(x_i|\boldsymbol{x}_{<i})}{|\boldsymbol{x}|}
\end{align}
\end{small}

\vspace{-1\baselineskip}
\noindent where $f(\boldsymbol{x})$ is the fluency score of sentence $\boldsymbol{x}$. $P(x_i|\boldsymbol{x}_{<i})$ is the probability of $x_i$ given context $\boldsymbol{x}_{<i}$, computed by a pre-trained language model, and $|\boldsymbol{x}|$ is the length of sentence $\boldsymbol{x}$.
\subsubsection*{FST: Formality discrimination}

For FST, we propose a novel feature augmentation method called \textit{formality discrimination}. The idea was already depicted in Figure \ref{example}, motivated by the observation that cross-lingual machine translation (MT) often changes the formality of a sentence.

We collect a number of informal English sentences from twitter and online forums, denoting as $\mathcal{S}=\{\bm{s}_i\}_{i=1}^{|\mathcal{S}|}$ where $\bm{s}_i$ denotes the $i$-th sentence. We first translate\footnote{https://translate.google.com/} them into a pivot language (e.g., Chinese) and then translate them back into English, as Figure \ref{example} shows. In this way, we obtain a rewritten sentence $\bm{s}_i'$ for each sentence $\bm{s}_i \in \mathcal{S}$.

To verify whether $\bm{s}_i'$ improves the formality of $\bm{s}_i$, we introduce a formality discriminator which is a binary classifier trained with formal text (e.g., news) and informal text (e.g., tweets) to quantify the formality level of a sentence. If the discriminator finds $\bm{s}_i'$ largely improves the formality of $\bm{s}_i$, then ($\bm{s}_i$, $\bm{s}_i'$) will be selected as augmented data:
\begin{equation}
\small
 \mathcal{T}_{aug} = \{(\bm{s}_i, \bm{s}_i') | P_{+}(\bm{s}_i') - P_{+}(\bm{s}_i) \ge \sigma \}
\end{equation}
where $P_{+}(\bm{x})$ is the probability of sentence $\bm{x}$ being formal, predicted by the discriminator, and $\sigma$ is the threshold for augmented data selection.

With this method, we can obtain much augmented data with valuable rewriting knowledge for FST that is not included by the original training data, which is helpful to generalize the model.

\subsubsection{Multi-task transfer}\label{subsubsec:multi-task}
In addition to back translation and feature discrimination that use artificially generated sentence pairs for data augmentation, we introduce multi-task transfer that uses annotated data from other seq2seq tasks, which may involve useful rewriting knowledge, as augmented data to benefit the target sentence rewriting task. A typical example is shown in Figure \ref{example}, in which GEC annotated data can provide knowledge to help the model correct grammatical errors in the input informal sentence for FST task.

For multi-task transfer, the augmented data from other tasks is supplementary and should not distract the model from the gold data. Fortunately, our pre-training \& fine-tuning approach allows the model to absorb useful knowledge from the augmented data without hurting the trust in the gold data. Therefore, we can introduce annotated data from other tasks that can potentially benefit the target rewriting task as augmented data to help pre-train the model.

\section{Experiments}\label{sec:experiment}
We use the Transformer \cite{vaswani2017attention} as our default seq2seq model and evaluate our approach in two important well-defined sentence rewriting tasks: \textbf{G}rammatical \textbf{E}rror \textbf{C}orrection (GEC) and \textbf{F}ormality \textbf{S}tyle \textbf{T}ransfer (FST), both of which have high-quality benchmark datasets with reliable references from multiple human annotators and evaluation metrics. 

\subsection{GEC Evaluation}\label{subsec:GEC_eval}
\subsubsection{Setting}

We test our approach on two well-known GEC benchmarks: CoNLL-2014 \cite{ng2014conll} and JFLEG \cite{napoles2017jfleg}. CoNLL-2014 contains 1,312 test sentences while JFLEG contains 747 test sentences. Being consistent with the official evaluation metrics, we use Max-Match (M$^2$) Precision, Recall and $F_{0.5}$ \cite{dahlmeier2012better} for CoNLL-2014 and GLEU \cite{napoles2015ground} for JFLEG evaluation. As previous studies, we use CoNLL-2013 test set and JFLEG dev set as our development sets for CoNLL-2014 and JFLEG respectively. As most of the systems \cite{sakaguchi2017grammatical,chollampatt2018multilayer,grundkiewicz2018near} that use an additional spell checker to resolve spelling errors in JFLEG, we follow \cite{ge2018fluency} to use the public Bing spell checker\footnote{https://azure.microsoft.com/en-us/services/cognitive-services/spell-check/} to fix spelling errors in JFLEG as preprocessing.


We follow the restricted setting where only the public resources can be used and use public Lang-8 \cite{mizumoto2011mining, tajiri2012tense} and NUCLE \cite{dahlmeier2013building} dataset as gold parallel data, as most previous work did. For data augmentation, we use a combination of back translation (Section \ref{subsubsec:bt}) and fluency discrimination (Section \ref{subsubsec:f-dis}) to generate 118M augmented data from English Wikipedia and the News Crawl corpus during 2007-2013. Specifically, for a correct sentence, a back translation model trained with the public GEC data first generates 10 best outputs; then a 5-gram language model \cite{junczys2016phrase} trained on Common Crawl works as the fluency discriminator to select one whose fluency score is lower than the correct sentence and pairs them as augmented data.

We use Transformer (big) in \newcite{vaswani2017attention} as our error correction model and back translation model, which has a 6-layer encoder and decoder with the dimensionality of 1,024 for both input and output and 4,096 for inner-layers, and 16 self-attention heads. We use a shared source-target vocabulary of 30,000 BPE \cite{DBLP:conf/acl/SennrichHB16a}
tokens and train the model on 8 Nvidia V100 GPUs, using Adam optimizer with $\beta_1$=0.9, $\beta_2$=0.98. We allow each batch to have at most 4,096 tokens per GPU. In pre-training, the learning rate is set to 0.0005 with warmup over the first 8,000 steps and then decreasing proportionally to the inverse square root of the number of steps, and dropout probability is set to 0.3; while in the fine-tuning phase, the learning rate is set to 0.0001 with warmup over the first 4,000 steps and inverse square root decay after warmup, and the dropout ratio is set to 0.2. We pre-train the model for 200k steps and fine-tune it up to 50k steps. For inference, we follow \newcite{chollampatt2018multilayer} to generate 12-best predictions and choose the best sentence after re-ranking with their edit operations and language model scores computed by the 5-gram Common Crawl language model.

\subsubsection{Results}

We compare our pre-training \& fine-tuning approach to simultaneous training with both gold and augmented data. According to Table \ref{tab:gec_comp}, the 118M augmented sentence pairs derived through back translation and fluency discrimination are less valuable than the gold pairs, only achieving 44.75 $F_{0.5}$ and 57.54 GLEU. Training with the gold data and the augmented data simultaneously does not bring large improvements, instead leads to a decrease of both precision and $F_{0.5}$ on the CoNLL-2014 test set. When we use up-sampling or down-sampling to balance the original data and augmented data, we see 1-3 absolute improvement in CoNLL and JFLEG. In contrast, our pre-training \& fine-tuning approach significantly improves the performance over the model trained with only original data, achieving 61.11 $F_{0.5}$ (+6.87 improvement) and 62.93 GLEU score (+2.77 improvement), which is much more than the improvements by the simultaneous training approaches.

Also, we confirm that fluency discrimination benefits GEC data augmentation by comparing the last two models in Table \ref{tab:gec_comp}, because it can help filter out unnecessary and undesirable edits, which makes the augmented data more informative and helpful in improving the performance.

\begin{table}[t]
\centering
\small
\scalebox{0.95}{
\begin{tabular}{l|ccc|c}
\hline
\multirow{2}{*}{\textbf{Model}} & \multicolumn{3}{c|}{\textbf{CoNLL-2014}} & \textbf{JFLEG} \\
 & \textbf{$P$} & \textbf{$R$} & \textbf{$F_{0.5}$} & \textbf{$GLEU$} \\ \hline
Original data & 60.15 & 38.94 & 54.24 & 60.16 \\ 
Augmented data & 48.25 & 34.68 & 44.75 & 57.54 \\ \hline
ST & 59.27 & 39.41 & 53.84 & 60.59 \\
ST (down-sampling) & 61.90 & 39.04 & 55.41 & 61.02 \\ 
ST (up-sampling) & 64.29 & 39.18 & 56.98 & 61.37 \\ \hline
PT\&FT & \bf 68.05 & \bf 43.40 & \bf 61.11 & \bf 62.93 \\
PT\&FT (w/o F-Dis) & 67.23 & 42.94 & 60.40 & 62.42 \\ 
\hline
\end{tabular}
}
\caption{The performance comparison of models trained with simultaneous training (ST) and our pre-training \& fine-tuning (PT\&FT) approach, and the ablation test for fluency discrimination (F-Dis). For ST, down-sampling and up-sampling are for balancing the size of the augmented data and the original data. Specifically, down-sampling samples augmented data to make it in the same size of the original data; while up-sampling increases the frequency of the original data so that it becomes in the same size with the augmented data.}
\label{tab:gec_comp}\vspace{-0.25cm}
\end{table}

\begin{table*}[t]
\centering
\small
\scalebox{0.95}{
\begin{tabular}{l|c|c|c} 

\hline
\multicolumn{1}{c|}{\multirow{2}{*}{\bf System}}  & \multirow{2}{*}{\bf Setting} & \bf CoNLL-2014  & \bf JFLEG \\ 
 &  & $F_{0.5}$  & $GLEU$ \\ \hline
No edit & - & -  & 40.54 \\ \hline
NUS18-CNN \cite{chollampatt2018multilayer} & R & 54.79 & 57.47 \\
NUS18-NeuQE \cite{chollampatt2018neural} & R & 56.52 & - \\
Adapted-transformer \cite{junczys2018approaching} & R  & 55.8 & 59.9 \\
SMT-NMT hybrid \cite{grundkiewicz2018near} & R & 56.25& 61.50 \\
Wiki edit + Round-trip translation \cite{Google_GEC} & R  & 60.4 & 63.3 \\ 
Copy-Augmented Transformer \cite{Yuanfudao_GEC} & R & 61.15 & 61.00 \\
\bf Our approach (R) & R  & \bf 62.61 & \bf 63.54 \\ \hline

Nested-RNN-seq2seq \cite{DBLP:conf/acl/JiWTGTG17} & U & 45.15 & 53.41 \\
Fluency Boost Learning \cite{ge2018reaching} & U & 61.34 & 61.41 \\ 
Wiki edit + Round-trip translation \cite{Google_GEC} & U  & 62.8 & 65.0 \\ 
\bf Our approach (U) & U  & \bf 66.77  & \bf 65.22 \\ \hline
\end{tabular}
}
\caption{Comparison to the state-of-the-art GEC systems. R denotes the restricted setting where only public GEC data can be used for training, while U denotes the unrestricted setting where any data can be used. 
}\label{tab:gec-sota}
\end{table*}

We compare our approach to the top-performing GEC systems\footnote{The results of some latest work (e.g., \cite{grundkiewicz2019neural}) using W\&I and LOCNESS corpus \cite{yannakoudakis2018developing} for training are not reported.} in CoNLL and JFLEG. In addition to the restricted setting in which only public GEC data can be used for training, we also evaluate our approach in the unrestricted setting in which any data can be used. In the unrestricted setting, we additionally include 1.4M Cambridge Learner Corpus \cite{nicholls2003cambridge} and 2.9M non-public Lang-8 data as gold data, as \newcite{ge2018reaching} did, and 85M sentence pairs augmented from English Gigaword using the same data augmentation methods as we used in the restricted setting, except that the back translation model is replaced with the one trained with both public and non-public GEC data. Like most state-of-the-art GEC systems, we train 4 models with different random initializations for ensemble decoding.

Table \ref{tab:gec-sota} shows the results evaluated in CoNLL and JFLEG benchmarks. According to Table \ref{tab:gec-sota}, our approach obtains the state-of-the-art results in both restricted and unrestricted settings. In the restricted setting, it achieves 62.61 $F_{0.5}$ in CoNLL-2014. In JFLEG, it achieves 63.54 GLEU score which is the new state-of-the-art result, even outperforming the multi-round decoding results of \newcite{grundkiewicz2018near} and \newcite{Google_GEC}. In the unrestricted setting, our approach significantly outperforms the previous state-of-the-art GEC systems, and achieves the best results for the GEC benchmarks by now.

\subsection{FST Evaluation}
Formality style transfer is a practical sentence rewriting task, aiming to paraphrase an input sentence into desired formality. In this paper, we focus on informal$\to$formal style transfer since it is more practical in real application scenarios.

\subsubsection{Setting}
We use GYAFC benchmark dataset \cite{DBLP:conf/naacl/RaoT18} for training and evaluation. GYAFC's training split contains a total of 110K annotated informal-formal parallel sentences, which are annotated via crowd-sourcing of two domains: \emph{Entertainment \& Music} (E\&M) and \emph{Family \& Relationships} (F\&R). In its test split, there are 1,146 and 1,332 informal sentences in E\&M and F\&R domain respectively and each informal sentence has 4 referential formal rewrites. Following prior work \cite{DBLP:conf/coling/NiuRC18}, we use GYAFC dev split as our development set and use tokenized BLEU as our automatic evaluation metric.

We use all the three data augmentation methods we introduced and obtain a total of 4.9M augmented parallel sentences. Among them, 1.6M are generated by back-translating formal sentences in E\&M and F\&R domain on Yahoo Answers L6 corpus, 1.5M are derived by formality discrimination (the threshold $\sigma$ = 0.5), and 1.8M are from the public GEC data (Lang-8 and NUCLE).

We use the Transformer (base) model in \newcite{vaswani2017attention} as the seq2seq model, which has 6-layer transformer blocks with embedding dimension of 512 for input and output and 2,048 for inner-layers, and 8 self-attention heads. We build a shared vocabulary of 20K BPE \cite{DBLP:conf/acl/SennrichHB16a} tokens, and adopt the Adam optimizer to train the model with batch size of 4,096 tokens per GPU, as in Section \ref{subsec:GEC_eval}. In pre-training, the dropout probability is set to 0.1, the learning rate is set to 0.0005 with 8000 warmup steps and scheduled to an inverse square root decay after warmup; while during fine-tuning, the learning rate is set to 0.00025. We pre-train the model for 80k steps and fine-tune the model for a total of 15k steps.

\subsubsection{Results}
Table~\ref{tab:fst_basic_result} shows results of the models trained with simultaneous training and our pre-training \& fine-tuning approach. As the results in GEC, simultaneously training with the augmented and original data leads to a performance decline, because the noisy augmented data cannot achieve desirable performance by itself and may hinder the model to learn from the gold data in simultaneous training. In contrast, PT\&FT only uses the augmented data in the pre-training phase and treats it as the prior knowledge which is supplementary to the gold training data, reducing the negative effects of the augmented data and improving the results.

\begin{table}[t]
	\centering
	\scalebox{0.8}{
	\begin{tabular}{l|c|c}
		\hline
		\multirow{2}{*}{\textbf{Model}} &\textbf{E\&M} &\textbf{F\&R} \\  
		&\textbf{$BLEU$}   &\textbf{$BLEU$} \\
		\hline
		Original data &69.44   &74.19  \\
        Augmented  data &51.83  &55.66 \\
        \hline
        ST &59.93  &63.16 \\
        ST (up-sampling)  &68.43 &73.04  \\
	    ST (down-sampling) &68.54 &73.69 \\
	    \hline
		PT\&FT &\textbf{72.63} &\textbf{77.01} \\
		\hline
	\end{tabular}}
	\caption{The comparison of simultaneous training (ST) and Pre-train \& Fine-tuning (PT\&FT) for FST. }\label{tab:fst_basic_result}
\end{table}

Table \ref{tab:fst-aug} compares the results of our pre-training \& fine-tuning approach with different data augmentation methods. Compared with back translation, the improvements of formality discrimination and multi-task transfer are more significant since they introduce new rewriting knowledge and valuable training signals. The combination of the augmented data further improves the performance, obtaining more than 2.5 absolute improvement over the baseline trained with only original data.

\begin{table}[t]
    \centering
	\scalebox{0.8}{
		\begin{tabular}{l|c|c}
			\hline
			\multirow{2}{*}{\textbf{Model}} &\textbf{E\&M} &\textbf{F\&R} \\  
			&\textbf{$BLEU$}   &\textbf{$BLEU$}   \\
			\hline
			Original data &69.44   &74.19  \\
            \hline
             \multicolumn{3}{c}{\textbf{Pre-training \& Fine-tuning}} \\
            \hline
		      + BT   &71.18   &75.34   \\
		    + F-Dis     &71.72    &76.24   \\
		    + M-Task    &71.91     &76.21  \\
			+ M-Task + F-Dis & 72.40 &76.92 \\
			+ BT + M-Task + F-Dis &\textbf{72.63}   &\textbf{77.01}   \\
			\hline
		\end{tabular}}\vspace{-0.1cm}
	\caption{The comparison of different data augmentation methods for FST.}\label{tab:fst-aug}
\end{table}

We compare our approach to the following previous approaches in GYAFC benchmarks:
\leftmargini=6mm
\begin{itemize}
    \item Rule, PBMT, NMT, PBMT-NMT: Rule-based, phrase-based MT, NMT, PBMT-NMT hybrid model in \newcite{DBLP:conf/naacl/RaoT18}.
    \item NMT-MTL: The state-of-the-art NMT model with multi-task learning \cite{DBLP:conf/coling/NiuRC18}.
\end{itemize}

According to the results in Table \ref{tab:fst-sota}, our single model outperforms the previous state-of-the-art ensemble model \cite{DBLP:conf/coling/NiuRC18} and our ensemble model achieves a new state-of-the-art result: 74.24 in E\&M and 77.97 in F\&R domain in GYAFC benchmark.
\begin{table}[t] 
    \centering
	\scalebox{0.8}{
		\begin{tabular}{c|c|c}
			\hline
			\multirow{2}{*}{\textbf{System}} &\textbf{E\&M} &\textbf{F\&R}  \\  
			&\textbf{$BLEU$}  &\textbf{$BLEU$}  \\
		    \hline
			No-edit &50.28 &51.67 \\
			\hline
			
			Rule &60.37 &66.40  \\
			PBMT &66.88 &72.40  \\
			NMT &58.27 &68.26 \\
			NMT-PBMT  &67.51  &73.78  \\
			NMT-MTL     &71.29 (72.01)  &74.51 (75.33)  \\
			\hline
            Our approach  &\textbf{72.63 (74.24)} &\textbf{77.01 (77.97)} \\
            \hline
		\end{tabular}}\vspace{-0.05cm}
	\caption{The comparison of our approach to the state-of-the-art result for FST. Numbers in parentheses are the results of ensemble of 4 models with different random initializations. }\label{tab:fst-sota}\vspace{-0.1cm}
\end{table}

\begin{table}[t]
\centering
	\scalebox{0.8}{
    \begin{tabular}{l|c|c|c}
    \hline
    Model & Formality  & Fluency & Meaning \\ \hline
    Original data &1.31~  &1.77~ &1.80~~ \\
    NMT-MTL &1.34~   &1.78~ &$\textbf{1.92}^*$ \\ \hline
   \textbf{Ours} &$\textbf{1.45}^*$ &$\textbf{1.85}^{*\dag}$  &$\textbf{1.92}^*$ \\
   \hline
    \end{tabular}}
    \caption{Results of human evaluation of FST. Scores marked with */{$^\dag$} are significantly different from the Original data/NMT-MTL scores ($p < 0.05$ in t-test).}
    \label{human_eval}\vspace{-0.2cm}
\end{table}

We also conduct human evaluation. Following previous work \cite{DBLP:conf/naacl/RaoT18}, we assess the model output on three criteria: \emph{formality}, \emph{fluency} and \emph{meaning preservation}. We compare our baseline model trained only with original data (in Table~\ref{tab:fst_basic_result}), the previous state-of-the-art model (NMT-MTL) and our PT\&FT approach. We randomly sample 300 items and each item includes an input and three corresponding outputs that shuffled to anonymize model identities. Two annotators are asked to rate these outputs on a discrete scale of 0 to 2.

Table~\ref{human_eval} presents the human evaluation results, showing that our model is consistently well rated in human evaluation. It significantly improves our baseline model trained with only original data in all three aspects, and outperforms the previous state-of-the-art model in terms of fluency ($p<0.05$ in t-test), confirming that our pre-training \& fine-tuning approach with data augmentation is helpful in improving FST task.

\vspace{-0.05cm}
\subsection{Discussion}
With the success of the pre-training \& fine-tuning approach in sentence rewriting, we study generalizing it to other seq2seq tasks. We use WMT14 English-German benchmark as our testbed and train and evaluate on the standard WMT14 English-German dataset. As previous work \cite{vaswani2017attention}, we validate on newstest2013. By removing the sentences longer than 250 words and sentence-pairs with a source/target length ratio exceeding 1.5 in training data, we obtain 3.9M parallel sentences as the original training data. For data augmentation, we back-translate 37M German mono-lingual sentences from News Crawl in 2013. 

We use the same model architecture and training configuration in Section \ref{subsec:GEC_eval} and compare the results of our pre-training \& fine-tuning approach and the simultaneous training approaches. Table 8 reports tokenized BLEU of our approach in the WMT14 English-German dataset. Our PT\&FT approach still outperforms the simultaneous training and achieves 32.2 BLEU. As far as we know, it is the best result for an MT model that uses only WMT14's mono- and bi-lingual data for training, which is only inferior to the commercial translation engine \textit{DeepL}\footnote{https://www.deepl.com/press.html} and \textit{FAIR}'s model trained with the larger WMT18 dataset (containing 5.2M bi-lingual sentence pairs) and 226M augmented sentence pairs through back translation simultaneously with up-sampling.

\begin{table}[t]
\small
\centering
\begin{tabular}{c|l|c} \hline
\multicolumn{2}{c|}{\bf Model} & \textbf{BLEU} \\ \hline
\multirow{2}{*}{SOTA} & \it DeepL & 33.3 \\
 & \textit{FAIR} \cite{edunov2018understanding} & 35.0 \\ \hline
\multirow{6}{*}{Ours} & Original data & 28.7 \\
 & Augmented data & 28.8 \\
 \cline{2-3}
 & ST & 29.3 \\
 & ST (up-sampling) & 31.3 \\
 & ST (down-sampling) & 31.0 \\
  \cline{2-3}
 & PT\&FT & 32.2 \\ \hline
\end{tabular}
\caption{Results in WMT14 English-German dataset.\label{tab:mt_sota}}
\end{table}

One interesting observation in Table \ref{tab:mt_sota} is that the augmented data itself can achieve the comparable performance to the original training data in MT. This is quite different from the results in the sentence rewriting tasks (i.e., GEC and FST) where the augmented data can only yield a low performance by itself. One reason is that in the MT experiment, the augmented data is in the same domain (i.e., news domain) with the test data; while in GEC and FST, the domain of augmented data is different from the test set. The other reason is that for many sentence rewriting tasks, most parts of a source sentence should not be edited unless necessary. Since the augmented data may contain various noisy and unnecessary editing signals, it is likely to make the model become aggressive to do erroneous rewrites, resulting in a low performance. Therefore, for sentence rewriting, the augmented data is better to be pre-trained than trained together with the original training data.

\vspace{-0.08cm}
\section{Related Work}

Pre-training approaches \cite{DBLP:conf/nips/DaiL15,DBLP:conf/emnlp/ConneauKSBB17,DBLP:conf/nips/McCannBXS17,DBLP:conf/acl/RuderH18} have drawn much attention recently. Among them, the most successful ones are ELMo \cite{DBLP:conf/naacl/PetersNIGCLZ18}, OpenAI-GPT \cite{GPT} and BERT \cite{DBLP:journals/corr/abs-1810-04805}, which are all based on pre-training a language model on massive unlabeled text data and fine-tuning with the task-specific gold data. While some previous work studies initializing a seq2seq model with a pre-trained language model \cite{DBLP:conf/emnlp/RamachandranLL17} and multi-task seq2seq learning \cite{luong2015multi}, there is no much work related to seq2seq pre-training with data augmentation until in the last few months when some work \cite{Google_GEC,Yuanfudao_GEC,grundkiewicz2019neural} have started to explore pre-training with augmented data for GEC.
Different from the studies that report better GEC performance through pre-training with augmented data, we focus on studying how to best utilize the augmented data by empirically comparing the effects of different training paradigms (i.e., \textit{simultaneous training} VS \textit{pre-training \& fine-tuning}) given the same augmented data in the final performance, analyzing the necessity of pre-training \& fine-tuning for seq2seq sentence rewriting tasks.

Our work is also related to the research exploring data augmentation methods in NLP. In addition to word substitution \cite{DBLP:conf/acl/FadaeeBM17a,zhou2019bert} and paraphrasing \cite{DBLP:conf/emnlp/DongMRL17}, back translation \cite{DBLP:conf/wmt/BojarT11,DBLP:conf/acl/SennrichHB16} including its variations \cite{he2016dual,zhang2018joint} attracts much attention as its success in MT \cite{DBLP:journals/corr/abs-1804-06189,edunov2018understanding}. 
For sentence rewriting, an important research branch for data augmentation is artificial error generation for GEC \cite{DBLP:conf/acl/BrockettDG06,DBLP:conf/bea/FosterA09,DBLP:conf/naacl/RozovskayaR10,DBLP:conf/acl/RozovskayaR11,DBLP:conf/bea/RozovskayaSR12,DBLP:conf/conll/FeliceYAYK14,DBLP:conf/bea/YuanBF16,DBLP:conf/bea/ReiFYB17,xie2018noising}, which studies generating source sentences with grammatical errors. Also, recent work uses back translation to obtain style-reduced paraphrases \cite{DBLP:conf/acl/TsvetkovBSP18} and employs the data from other tasks with the same target language to enhance the model in terms of target language modeling for sentence rewriting tasks \cite{DBLP:conf/coling/NiuRC18}.

\section{Conclusion and Future Work}

In this paper, we study seq2seq pre-training \& fine-tuning with various data augmentation methods in sentence rewriting. Extensive experiments demonstrate that our proposed data augmentation methods can effectively improve the performance and that pre-training \& fine-tuning with data augmentation has advantages over the conventional simultaneous training approaches. It achieves new state-of-the-art results in multiple benchmarks in GEC and FST sentence rewriting tasks. In the future, we plan to generalize the current task-specific seq2seq pre-training approach so that we could pre-train a task-independent seq2seq model as a base for any monolingual sentence rewriting task.

\bibliography{emnlp-ijcnlp-2019}
\bibliographystyle{acl_natbib}

\end{document}